\documentclass{article}

\usepackage[preprint]{neurips_2024}

\usepackage[utf8]{inputenc} 
\usepackage[T1]{fontenc}    
\usepackage{hyperref}       
\usepackage{url}            
\usepackage{booktabs}       
\usepackage{amsfonts}       
\usepackage{nicefrac}       
\usepackage{microtype}      
\usepackage{amsmath}
\usepackage{graphicx}
\usepackage{multirow}
\usepackage{diagbox}
\usepackage[table,xcdraw]{xcolor}
\usepackage{CJKutf8}
\usepackage{algorithm,algpseudocode}

\title{BiSup: Bidirectional Quantization Error Suppression for Large Language Models}

\author{
  Minghui Zou \\
  College of Intelligence and Computing \\
  Tianjin University \\
  \texttt{minghuizou@tju.edu.cn} \\
  \And
  Ronghui Guo \\
  College of Intelligence and Computing \\
  Tianjin University \\
  \texttt{ronghui\_guo@tju.edu.cn} \\
  \And
  Sai Zhang \\
  College of Intelligence and Computing \\
  Tianjin University \\
  \texttt{zhang\_sai@tju.edu.cn} \\
  \And
  Xiaowang Zhang \\
  College of Intelligence and Computing \\
  Tianjin University \\
  \texttt{xiaowangzhang@tju.edu.cn} \\
  \And
  Zhiyong Feng\thanks{Corresponding Author} \\
  College of Intelligence and Computing \\
  Tianjin University \\
  \texttt{zyfeng@tju.edu.cn} \\
}

\begin{document}

\begin{CJK}{UTF8}{gbsn}

\maketitle


\begin{abstract}

As the size and context length of Large Language Models (LLMs) grow, weight-activation quantization has emerged as a crucial technique for efficient deployment of LLMs. Compared to weight-only quantization, weight-activation quantization presents greater challenges due to the presence of outliers in activations. Existing methods have made significant progress by exploring mixed-precision quantization and outlier suppression. However, these methods primarily focus on optimizing the results of single matrix multiplication, neglecting the bidirectional propagation of quantization errors in LLMs. Specifically, errors accumulate vertically within the same token through layers, and diffuse horizontally across different tokens due to self-attention mechanisms. To address this issue, we introduce BiSup, a \textbf{Bi}directional quantization error \textbf{Sup}pression method. By constructing appropriate optimizable parameter spaces, BiSup utilizes a small amount of data for quantization-aware parameter-efficient fine-tuning to suppress the error vertical accumulation. Besides, BiSup employs prompt mixed-precision quantization strategy, which preserves high precision for the key-value cache of system prompts, to mitigate the error horizontal diffusion. Extensive experiments on Llama and Qwen families demonstrate that BiSup can improve performance over two state-of-the-art methods (the average WikiText2 perplexity decreases from 13.26 to 9.41 for Atom and from 14.33 to 7.85 for QuaRot under the W3A3-g128 configuration), further facilitating the practical applications of low-bit weight-activation quantization.

\end{abstract}


\section{Introduction}

The emergence and development of large language models (LLMs) have had a disruptive impact across various domains. Empirical evidence indicates that, compared to previous pre-trained language models (PLMs), LLMs can demonstrate powerful emergent capabilities when their scale reaches a certain threshold, thereby better adapting to complex real-world applications \citep{zhao2023survey}. However, the growth in model size is accompanied by an increase in the computational resources required for model deployment and inference, limiting the widespread application of LLMs in various resource-constrained scenarios, such as the edge.


To reduce the deployment and inference costs of models, model quantization has emerged as a promising approach \citep{nagel2021white,zhu2023survey,wang2024model}. LLMs are typically trained and stored at high precision (e.g., FP16 or BF16). The objective of model quantization is to convert them into lower precision (e.g., INT4), thereby significantly reducing memory consumption while leveraging hardware features to accelerate inference speed. Model quantization comprises two main directions: Quantization-Aware Training (QAT) and Post-Training Quantization (PTQ). Compared to QAT methods, which require substantial data and computational resources, PTQ methods are widely used in practical applications owing to their low cost and high yield. From the perspective of quanitized objects, model quantization encompasses two primary branches: weight-only quantization and weight-activation quantization. The weight-only quantization methods have been applied in various inference frameworks \citep{frantar2022optq,lin2023awq}. Nonetheless, as the model sizes and context lengths expand, the proportion of memory consumption attributed to activations escalates, thus rendering activation quantization a critical research issue.


Compared to weight-only quantization, weight-activation quantization poses greater challenges due to the presence of outliers in activations. A recognized finding is that while outliers are difficult to be quantized, they typically occur only in a few specific channels \citep{dettmers2022gpt3}. Accordingly, existing methods are mainly investigated in two directions: mixed-precision quantization \citep{dettmers2022gpt3,guo2023olive,ashkboos2023quik} and outlier suppression \citep{xiao2023smoothquant,shao2023omniquant,ma2024affinequant}. The core idea of mixed-precision quantization methods lies in retaining a minority of outlier channels at high precision while quantizing the majority of normal channels to low precision. LLM.int8() \citep{dettmers2022gpt3} identifies outlier channels based on the magnitude of activations and then decomposes the weight matrix and activation matrix by channel. While the outlier matrices are multiplied at high precision, the normal matrices are first quantized to low precision. Building upon LLM.int8(), QUIK \citep{ashkboos2023quik} proposes improvements by replacing the RTN algorithm with the GPTQ \citep{frantar2022optq} and moving high-precision outlier channels to the end to compensate for quantization losses. Outlier suppression methods are grounded on the premise that, compared to activations, weights exhibit a smoother distribution and better quantization properties. Therefore, by means of equivalent transformations, outliers in activations can be transferred to weights to reduce the difficulty of activation quantization. SmoothQuant \citep{xiao2023smoothquant} proposes an empirical formula to calculate a scaling factor based on the maximum values of activations and weights. OmniQuant \citep{shao2023omniquant} suggests that manually designed rules for offsetting and scaling outlier channels may not achieve optimal results, thus using gradient optimization methods as an alternative. AffineQuant \citep{ma2024affinequant} proposes an equivalent affine transformation, unifying previous equivalent transformation methods, and introduces a new optimization algorithm to ensure the reversibility of the transformation matrix.


\begin{figure}[ht]
  \centering
  \includegraphics[width=\textwidth]{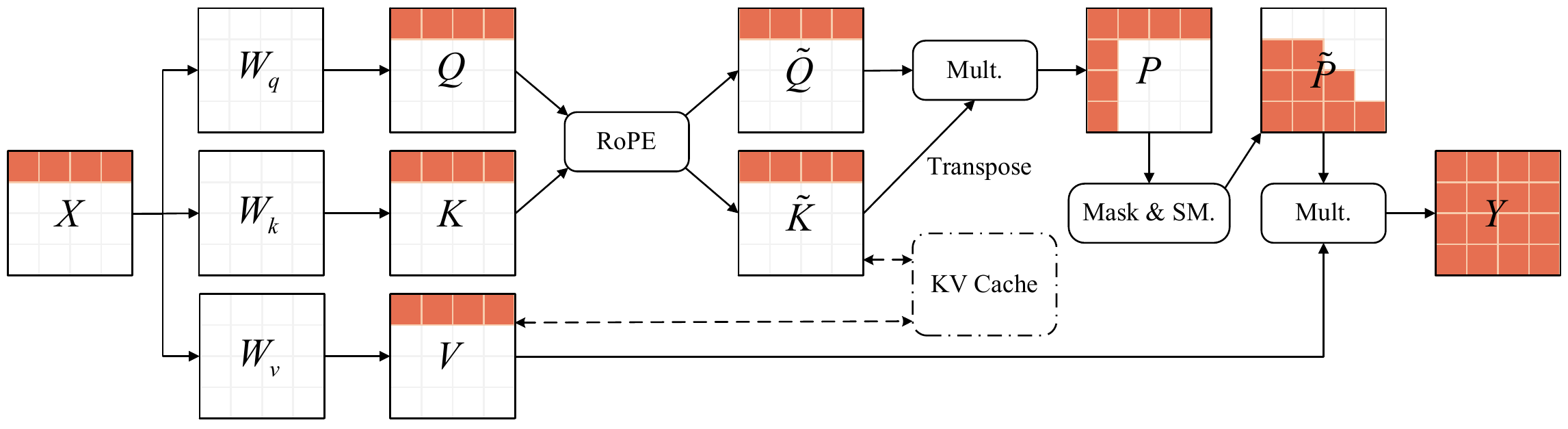}
  \caption{The error propagation within attention block under activation-only quantization. Different colors on the tensor represent different meanings, where orange indicates that it contains quantization error and white the opposite.}
  \label{fig:propagation}
\end{figure}
While existing methods have made significant progress in mitigating the effects of outliers, they primarily focus on optimizing the results of single matrix multiplication, neglecting the bidirectional propagation of quantization errors in the model. This includes the error vertical accumulation within the same token through layers and the error horizontal diffusion between different tokens due to self-attention mechanisms. For convenience, we consider the error propagation within attention block under activation-only quantization, as illustrated in Figure \ref{fig:propagation}. Within the block, there are consecutive matrix multiplications necessitating corresponding quantization and dequantization operations. The error introduced by the preceding quantized matrix multiplication influences the subsequent one, leading to error accumulation. Besides, because self-attention mechanisms involve interactions between different tokens, quantization errors appearing on one token will diffuse to other tokens, further contributing to a wider range of error vertical accumulation and resulting in a vicious cycle. Consequently, effectively suppressing bidirectional error propagation inside the model is a crucial step to ensure the effectiveness of existing high-performance quantization methods.


In this paper, we introduce BiSup, a \textbf{Bi}directional quantization error \textbf{Sup}pression method. To alleviate the error vertical accumulation, it is required to minimize the quantization error of single matrix multiplication while counteracting the existing quantization error in time. Inspired by OmniQuant \citep{shao2023omniquant} and QLoRA \citep{dettmers2024qlora,liao2024apiq}, BiSup starts from the weight-activation quantization formula, considers the distribution pattern of activation outliers and the convergence of few-shot fine-tuning to design optimizable parameter spaces that require only a small amount of data and computational resources for calibration. To mitigate error horizontal diffusion, it is advantageous to preserve the accuracy of important tokens (i.e., tokens with larger attention weights). Analyzing the attention maps of LLMs, we notice that LLMs have a strong dependence on the first token. From the perspective of system prompt caching, BiSup introduces the prompt mixed-precision quantization strategy. By keeping a small amount of system prompt cache at high precision, the error of important token interactions can be effectively minimized. Our contributions are as follows:
\begin{itemize}
    \item We indicate the bidirectional propagation of quantization errors, providing a new research perspective for post-training quantization and further improving the practical value of existing quantization methods.
    \item For the error vertical accumulation, we design appropriate parameter spaces to employ quantization-aware parameter-efficient fine-tuning, integrating the advantages of post-training quantization and quantization-aware training methods.
    \item For the error horizontal diffusion, we propose the prompt mixed-precision quantization strategy that ensures the accuracy of important token interactions by retaining a small number of high-precision system prompts cache.
    \item Extensive experiments on Llama and Qwen families show that BiSup can further improve performance on the top of two state-of-the-art methods (the average WikiText2 perplexity decreases from 13.26 to 9.41 for Atom and from 14.33 to 7.85 for QuaRot under the W3A3-g128 configuration).
\end{itemize}


\section{Related Work}


\paragraph{Quantization-Aware Training (QAT)} QAT methods \citep{liu2023llm,dettmers2024qlora,li2023loftq} require substantial data and computational resources to fine-tune the quantized LLMs. To address the challenge of acquiring fine-tuning data, LLM-QAT \citep{liu2023llm} proposes a data-free knowledge distillation method that first utilizes LLMs themselves to generate a large amount of data and then distills the quanitized model. To reduce memory consumption during fine-tuning on downstream tasks while addressing the discrepancy between training and inference (i.e., quantization awareness), QLoRA \citep{dettmers2024qlora} combines LoRA \citep{hu2021lora} with weight-only quantization, significantly reducing the cost of fine-tuning. Furthermore, LoftQ \citep{li2023loftq} indicates the shortcomings of the parameter initialization method used in QLoRA and employs Singular Value Decomposition (SVD) to determine the appropriate low-rank initialization parameters. This initialization technique significantly enhances the generalizability of quantized models.


\paragraph{Post-Training Quantization (PTQ)} PTQ methods \citep{frantar2022optq,shao2023omniquant,li2024norm,ding2023cbq} generally require only a small amount of calibration data and computational resources to calibrate the quantized models. One of the most widely used PTQ methods is GPTQ \citep{frantar2022optq} , which is compatible with most existing quantization methods and often serves as an enhanced alternative to the RTN algorithm. GPTQ performs per-channel quantization of all parameters within a block, and after quantizing each channel's parameters, it appropriately adjusts the remaining unquantized parameters within the block to compensate for the accuracy loss. Recently, given the distinct advantages of QAT and PTQ (where QAT avoids complex algorithm design through gradient optimization and PTQ offers high-yield at low-cost), some studies \citep{shao2023omniquant,li2024norm,ding2023cbq} have explored combining these two approaches which utilize a small amount of data to optimize a carefully designed parameter space. OmniQuant \citep{shao2023omniquant} and CBQ \citep{ding2023cbq} learn appropriate truncation thresholds for weights and smoothing factors for outliers in activations. Norm Tweaking \citep{li2024norm} restricts the trainable parameter space to the weights in the LayerNorm and performs cross-block optimization. Our proposed BiSup makes adaptive improvements based on previous studies by considering weight-activation quantization settings thoroughly.



\section{Preliminaries}
Quantization techniques involve converting high-precision floating-point numbers into lower-precision representations to reduce the memory and computation requirements of models while maintaining their performance. RTN (round-to-nearest) algorithm is widely utilized due to its simplicity and hardware efficiency, encompassing both asymmetric and symmetric quantization schemas \citep{nagel2021white}. While asymmetric quantization typically offers better performance, symmetric quantization is generally favored in weight-activation quantization settings due to its lower computation and implementation complexities. The quantization process involves two main steps: first calculating the quantization parameters (usually including scale factor and zero point), and then quantizing the input tensor. The quantization parameters (scale factor only since zero point is always equal to 0) for symmetric quantization are determined by \citep{jacob2018quantization}:
\begin{equation}\label{eq:scale}
    \Delta=\frac{\max(|X|)}{2^{N-1}-1},
\end{equation}
where $X$ represents the input tensor to be quantized, $N$ denotes the quantization bit-width, and $\Delta$ signifies the scale factor. Subsequently, the input tensor is quantized as follows:
\begin{equation}\label{eq:rtn}
    \bar{X}=\mathrm{clamp}(\lfloor\frac{X}{\Delta}\rceil,-2^{N-1},2^{N-1}-1),
\end{equation}
where $\bar{X}$ is the quantized tensor, $\lfloor\cdot\rceil$ denotes the rounding operation, and $\mathrm{clamp}(\cdot)$ represents the truncation operation.


Due to the presence of outliers in the input tensor, applying the same set of quantization parameters to the entire tensor (\textbf{tensor-wise}) may lead to degradation, hence quantization is often performed at a finer granularity (\textbf{channel/token-wise}), which computing a set of quantization parameters for each row or column in the tensor. Furthermore, each channel/token can be subdivided into multiple groups and quantized within the groups (\textbf{group-wise}), which is the most widely used setting in LLMs quantization.


\section{Methodology}

In this section, we describe the proposed bidirectional quantization error suppression method (BiSup). For the error vertical accumulation, BiSup starts from the weight-activation quantization formula (\textsection \ref{sec:clipping}), considers the distribution pattern of activation outliers (\textsection \ref{sec:smoothing}) and the convergence of few-shot fine-tuning (\textsection \ref{sec:lorc}) to design parameter spaces that can be optimized by gradient descent methods (\textsection \ref{sec:peft}). For the error horizontal diffusion, by analyzing the distribution of attention weights, we show that LLMs have a strong dependence on the first token. From the perspective of system prompt caching, BiSup introduces the prompt mixed-precision quantization strategy (\textsection \ref{sec:prompt}) that effectively reduces the error of important token interactions. The overall algorithm is presented in \textsection \ref{apx:overall algorithm}.

\subsection{Fine-Grained Weight-Activation Clipping}
\label{sec:clipping}

Analyzing the equations (\ref{eq:scale}) and (\ref{eq:rtn}), it can be seen that although the entire quantization interval contains $2^N$ integers, the number of actual valid integers is only $2^N-1$ (where $-2^{N-1}$ is never taken). When $N$ is small, the resulting bit-width waste is non-negligible. Fortunately, the clipping mechanism adopted in existing methods \citep{shao2023omniquant,zhao2023atom,ashkboos2024quarot} from the perspective of rounding error can effectively solve this problem, which allows the quantized values to be distributed over the whole interval by truncating the maximum and minimum value. Note that the bit-width waste due to symmetric quantization is also present in activation quantization, so it is necessary to adopt clipping mechanism in both weight and activation quantization, which is often overlooked. The scale factor with the introduction of the clipping mechanism is determined by:
\begin{equation}\label{eq:scale clip}
    \Delta_g=\frac{\max(|X_g|)}{2^{N-1}-1}\times c,
\end{equation}
where $g$ is the group index of group-wise quantization and $c$ represents the clip value, which is typically obtained through grid search. However, using the same clip value $c$ for the entire tensor, without considering distribution differences between different groups, leads to sub-optimal results. Therefore, we propose fine-grained weight-activation clippping, where clip values $c_g$ are computed individually for each group and set as a learnable parameter, so that it can be optimized using gradient descent methods along with other parameters instead of grid search. Here, we obtain the optimizable parameter space $\Theta_1=\{c_g\}$.

\subsection{Soft-Constrained Weight-Activation Smoothing}
\label{sec:smoothing}

As stated above, it can be observed that activation is more difficult to be quantized due to outliers, while outliers tend to be concentrated in some specific channels. Consequently, some works \citep{xiao2023smoothquant,shao2023omniquant,ma2024affinequant} have employed the smoothing mechanism to alleviate this problem, with the core idea being to migrate the difficulty of activation quantization to weight quantization. Specifically, activations are scaled down to smooth the distribution, while weights need to be scaled up to ensure computational invariance. This process can be defined as follows:
\begin{equation}\label{eq:smoothing}
    \langle X \rangle \cdot \langle W \rangle \approx \langle X \mathrm{diag}(s)^{-1} \rangle 
 \cdot\langle \mathrm{diag}(s)W \rangle,
\end{equation}
where $s$ denotes the smoothing factor, with dimensions equal to the number of columns of $X$ (i.e., the output dimension), and $\langle\cdot \rangle$ represents quantization operation.


To obtain the smoothing factor, SmoothQuant \citep{xiao2023smoothquant} designs an empirical formula, while OmniQuant \citep{shao2023omniquant} represents it as a learnable parameter. Here, we use the same approach as OmniQuant with improvements. In the original smoothing mechanism, to ensure computational invariance, the transformation matrix applied to activation and weight must satisfy a reversible relationship. However, this constraint is unnecessary in training-based methods since the optimization objective is already able to guarantee invariance. Therefore, we propose soft-constrained weight-activation smoothing to extend flexibility, formalized as:
\begin{equation}\label{eq:soft-constrained smoothing}
    \langle X \rangle \cdot \langle W \rangle \approx \langle X \mathrm{diag}(s_1) \rangle 
 \cdot\langle \mathrm{diag}(s_2)W \rangle,
\end{equation}
where $s_1$ and $s_2$ respectively represent the smoothing factors applied to activation and weight, with the constraint $\mathrm{diag}(s_1)\cdot\mathrm{diag}(s_2)=I$ ensured during parameter initialization. Noting that the optimized $s_1$ can be integrated into the weight of RMSNorm \citep{zhang2019root}, and $s_2$ can be fused into $W$ directly. Thus, the smoothing mechanism incurs no additional overhead during the inference stage. Here, we obtain the optimizable parameter space $\Theta_2=\{s_1, s_2\}$.

\subsection{Stabilized Low-Rank Error Compensation}
\label{sec:lorc}

To compensate quantization errors, ZeroQuant-V2 \citep{yao2024exploring} proposes Low-Rank Compensation (LoRC). Given a weight matrix $W$ and quantized counterpart $\bar{W}$, the weight quantization error $E=W-\bar{W}$. LoRC employs two low-rank matrices $\hat{U}$ and $\hat{V}$ to estimate $\hat{E}=\hat{U}\hat{V}$, where $\hat{U}$ and $\hat{V}$ are obtained by performing Singular Value Decomposition (SVD) on $E$ and selecting the top $r$ singular values. Although LoRC plays well, it requires to retain two additional low-rank matrices and introduces extra computational overhead during inference (i.e., $\hat{Y}=\bar{X}\bar{W}+(\bar{X}\hat{U})\hat{V}$). Furthermore, LoRC falls under the training-free methods, focusing primarily on compensating for weight quantization errors, and struggles to effectively reduce activation quantization errors.
Inspired by parameter-efficient fine-tuning methods \citep{dettmers2024qlora,liao2024apiq}, we enhance LoRC by making the two low-rank matrices trainable parameters, denoted as $A$ and $B$, which can be fused into $W$ after optimization. The optimization objective can be formalized as:
\begin{equation}\label{eq:low-rank error compensation}
    \underset{A,B}{\mathrm{argmin}}\|XW-\langle X \rangle \langle W+AB \rangle \|_F,
\end{equation}
where $W\in\mathbb{R}^{d_1\times d_2}$, $A\in\mathbb{R}^{d_1\times r}$, $B\in\mathbb{R}^{r\times d_2}$, and $r$ is a fixed hyperparameter. Equation (\ref{eq:low-rank error compensation}) has a similar form to LoRA \citep{hu2021lora} that is widely used. However, as revealed by comparative experiments (\textsection \ref{apx:comparative experiments}), directly applying it to post-training quantization (usually using a few calibration samples and having a small loss) may be difficult to converge. Therefore, we make an adjustment to Equation (\ref{eq:low-rank error compensation}):
\begin{equation}\label{eq:stabilized low-rank error compensation}
    \underset{A,B}{\mathrm{argmin}}\|XW-\langle X \rangle \langle W\cdot(1+AB) \rangle \|_F,
\end{equation}
where $W$ plays the role of anchor to enhance the stability of the optimization. Here, we obtain the optimizable parameter space $\Theta_3=\{A, B\}$.

\subsection{Quantization-Aware Parameter-Efficient Fine-Tuning}
\label{sec:peft}

In summary, in order to suppress the error vertical accumulation, we construct an optimizable parameter space $\Theta=\{\Theta_1, \Theta_2, \Theta_3\}$. Making a compromise between accuracy and memory consumption, similar with previous work \citep{shao2023omniquant,li2024norm}, we adopt a layer-wise optimization strategy. The optimization objective can be formalized as:
\begin{equation}\label{eq:loss}
\underset{\Theta}{\mathrm{argmin}} \| \mathcal{F}(X,W)-\mathcal{F}(\langle \bar X;\Theta_1,\Theta_2 \rangle, \langle W;\Theta_1,\Theta_2,\Theta_3 \rangle) \|_F,
\end{equation}
where $\mathcal{F}$ denotes a transformer layer, $\bar X$ represents the quantized activation produced by the previous layer. Note that we use the quantized activation as the input to the subsequent quantized layer, which enables us to timely eliminate quantization errors. The detailed procedure is shown in Algorithm \ref{alg:bisup}.

\subsection{Prompt Mixed-Precision Quantization}
\label{sec:prompt}

\begin{figure}[ht]
  \centering
  \includegraphics[width=\textwidth]{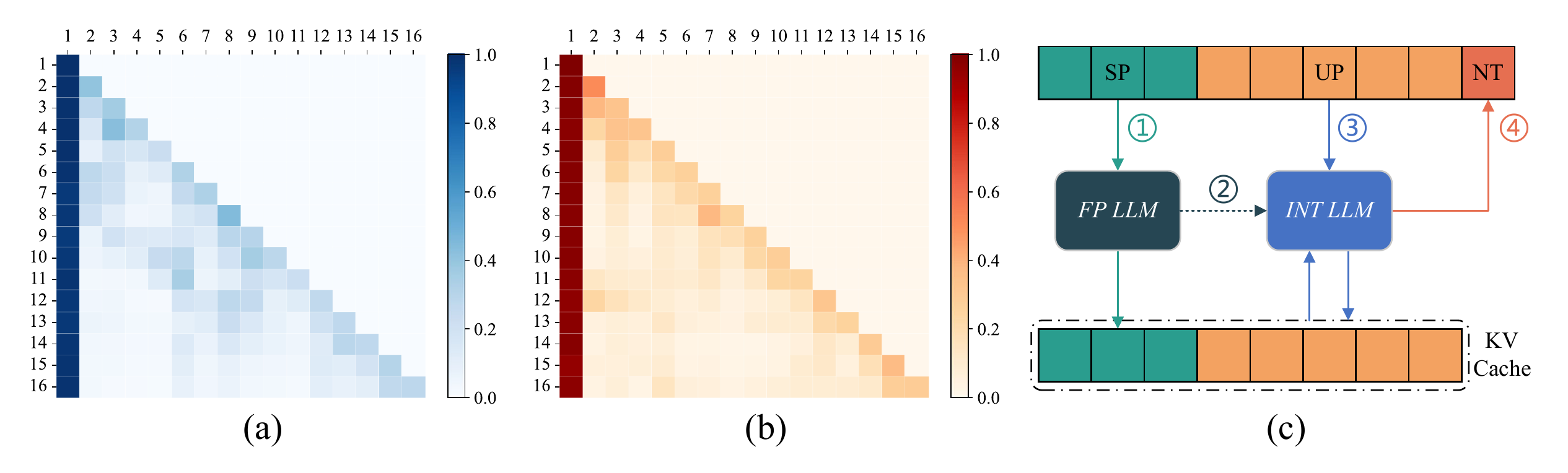}
  \caption{(a) and (b) show the attention maps of Llama3-8B and Llama3-8B-Instruct, respectively, and (c) illustrates the flow of the prompt mixed-precision quantization strategy. The symbols in (c) are explained as follows: SP, UP, and NT represent the system prompt, user prompt, and newly generated next token, respectively. These three elements together constitute the input to the LLM. \textit{FP LLM} refers to the original LLM, while \textit{INT LLM} denotes the quantized counterpart. Step ① indicates encoding and storing the system prompt into the KV cache at high precision. Step ② involves the model quantization. Step ③ describes the inference of the user prompt, which includes interactions with the mixed-precision KV cache. Step ④ denotes the prediction of the next token.}
  \label{fig:distribution}
\end{figure}
Due to the diffusion of quantization errors through the attention mechanism, it is necessary to analyze the interaction patterns between different tokens within the model. Figure \ref{fig:distribution} presents the attention maps of Llama3-8B \footnote{\url{https://huggingface.co/meta-llama/Meta-Llama-3-8B}} and Llama3-8B-Instruct \footnote{\url{https://huggingface.co/meta-llama/Meta-Llama-3-8B-Instruct}}. It can be observed that, regardless of whether undergoing instruction tuning \citep{zhang2023instruction}, the model exhibits a strong dependency on the first token. Therefore, preserving the accuracy of the first token's representation contributes to maintaining the interaction patterns among tokens, thereby suppressing error diffusion. More generally, the first token in instruction-tuned LLMs constitutes a part of the system prompt. In system implementation, these fixed special prompts are pre-computed and stored in a key-value cache for reuse across different requests \citep{kwon2023efficient}. Hence, we propose prompt mixed-precision quantization strategy, effectively reducing errors in the interactions with important tokens by retaining a few system prompt at high precision and quantize the user prompt normally. The detailed process is depicted in Figure \ref{fig:distribution}.


\section{Experiments}

We design experiments to address the following research questions (RQs):
\begin{itemize}
\item \textbf{RQ1:} Can BiSup achieve consistent performance improvements under various quantization configurations, and do all of the proposed techniques work?
\item \textbf{RQ2:} Does BiSup effectively suppress the error bidirectional propagation?
\item \textbf{RQ3:} What are the limitations of BiSup and what opportunities exist for its improvement?
\end{itemize}
To address RQ1, we evaluate the quantized models with language generation tasks and zero-shot tasks under three configurations (\textsection \ref{sec:overall results}) and conduct ablation studies (\textsection \ref{sec:ablation studies}). To tackle RQ2, we compare and analyze the quantization errors of different methods (\textsection \ref{sec:error suppression analysis}). Finally, we discuss the limitations and improvements of BiSup to answer RQ3 (\textsection \ref{sec:limitations and improvements}).

\subsection{Settings}
\label{sec:main settings}
\paragraph{Setup} We conduct all experiments on RTX 4090 using Hugging Face \citep{wolf2020transformers} and PyTorch \citep{paszke2019pytorch}. We apply INT4/INT3 per-channel/token symmetric quantization to weight/activation and per-token asymmetric quantization to key-value cache. Group-wise quantization is represented by ‘g’ (e.g., W4A4-g128 means 4-bit weight/activation quantization with a 128-group size). In our experiments, we always keep the quantization bit-width of the key-value cache consistent with the activation, and the weight and activation have the same group size while the key-value cache is not grouped. Consistent with previous works \citep{zhao2023atom,ashkboos2024quarot}, all intermediate activations are quantized except for Q and attention weights. We use grid search to obtain the optimal weight clip value and fix the activation clip value to 0.9 for baselines. We randomly selected 128 samples of length 2048 from WikiText2 \citep{merity2016pointer} as the calibration dataset. The $r$ used in stabilized low-rank error compensation is 32. We utilize the AdamW optimizer with a learning rate of 0.005, an epoch of 5, and a batch size of 8. To ensure fairness, we reproduce the baselines to unify all quantification settings and eliminate the effect of package versions on the experimental results.

\paragraph{Models and Tasks} We evaluate on Llama \citep{touvron2023llama} and Qwen \citep{qwen} families, which are limited to models up to 32B due to computational constraints. Besides, we also evaluate on the instruction-tuned Llama3-8B-Instruct to verify the generalizability. We report the perplexity of two language generation tasks (including WikiText2 \citep{merity2016pointer} and C4 \citep{raffel2020exploring}) and the accuracy of six zero-shot tasks (including ARC \citep{clark2018think}, BoolQ \citep{clark2019boolq}, HellaSwag \citep{zellers2019hellaswag}, PIQA \citep{bisk2020piqa}, and Winogrande \citep{sakaguchi2021winogrande}).

\paragraph{Baselines} Weight-activation quantization primarily involves two directions: mixed precision quantization and outlier suppression. We chose the state-of-the-art (SOTA) approaches from both directions for comparison. Atom \citep{zhao2023atom} is the SOTA of mixed-precision quantization that retains 128 outlier channels and employs fine-grained per-group quantization (g128). It is notable that in our experiments, Atom is same as QUIK \citep{ashkboos2023quik} when group-wise quantization is not used (i.e., W4A4). QuaRot \citep{ashkboos2024quarot} is the SOTA of outlier suppression that introduces offline and online Hadamard transformations to significantly improve outliers in activations at the cost of additional transform overhead. In this paper, Atom\_BiSup (or QuaRot\_BiSup) denotes using BiSup to suppress quantization errors of the model quantized by Atom (or QuaRot). Note that we do not evaluate QuaRot and QuaRot\_BiSup on Qwen family, which would crash due to numerical error accumulation.

\subsection{Overall Results}
\label{sec:overall results}

\begin{table}[ht]
\centering
\caption{WikiText2 perplexity results of Llama family. C4 perplexity results of Llama family can be found in Table \ref{tab:llama c4 ppl}. More perplexity results of Qwen family are in \textsection \ref{apx:additional overall results}.}
\label{tab:llama wiki ppl}
\begin{tabular}{cccccccc}
\hline
\multicolumn{2}{c}{\textbf{Llama1\&2\&3 / PPL$\downarrow$}}        & \textbf{1-7B}                         & \textbf{1-13B}                        & \textbf{1-30B}                        & \textbf{2-7B}                         & \textbf{2-13B}                        & \textbf{3-8B}                          \\ \hline
FP16                        & -                                      & 5.68                                  & 5.08                                  & 4.09                                  & 5.48                                  & 4.88                                  & 6.14                                   \\ \hline
                            & Atom                                   & 9.37                                  & 8.46                                  & 7.44                                  & 10.03                                 & 8.46                                  & 43.60                                  \\
                            & \cellcolor[HTML]{E7E6E6}Atom\_BiSup   & \cellcolor[HTML]{E7E6E6}\textbf{7.73} & \cellcolor[HTML]{E7E6E6}\textbf{6.98} & \cellcolor[HTML]{E7E6E6}\textbf{6.12} & \cellcolor[HTML]{E7E6E6}\textbf{7.98} & \cellcolor[HTML]{E7E6E6}\textbf{7.11} & \cellcolor[HTML]{E7E6E6}\textbf{16.32} \\
                            & QuaRot                                 & 6.27                                  & 5.46                                  & 4.58                                  & 6.16                                  & 5.38                                  & 8.22                                   \\
\multirow{-4}{*}{W4A4}   & \cellcolor[HTML]{E7E6E6}QuaRot\_BiSup & \cellcolor[HTML]{E7E6E6}\textbf{6.01} & \cellcolor[HTML]{E7E6E6}\textbf{5.36} & \cellcolor[HTML]{E7E6E6}\textbf{4.41} & \cellcolor[HTML]{E7E6E6}\textbf{5.88} & \cellcolor[HTML]{E7E6E6}\textbf{5.16} & \cellcolor[HTML]{E7E6E6}\textbf{7.42}  \\ \hline
                            & Atom                                   & \textbf{6.15}                         & \textbf{5.43}                         & \textbf{4.51}                         & 6.02                                  & \textbf{5.25}                         & \textbf{7.45}                          \\
                            & \cellcolor[HTML]{E7E6E6}Atom\_BiSup   & \cellcolor[HTML]{E7E6E6}6.15          & \cellcolor[HTML]{E7E6E6}5.45          & \cellcolor[HTML]{E7E6E6}4.53          & \cellcolor[HTML]{E7E6E6}\textbf{5.97} & \cellcolor[HTML]{E7E6E6}5.26          & \cellcolor[HTML]{E7E6E6}7.55           \\
                            & QuaRot                                 & 6.06                                  & 5.40                                  & 4.41                                  & 5.93                                  & 5.25                                  & 7.33                                   \\
\multirow{-4}{*}{W4A4-g128} & \cellcolor[HTML]{E7E6E6}QuaRot\_BiSup & \cellcolor[HTML]{E7E6E6}\textbf{5.97} & \cellcolor[HTML]{E7E6E6}\textbf{5.31} & \cellcolor[HTML]{E7E6E6}\textbf{4.37} & \cellcolor[HTML]{E7E6E6}\textbf{5.80} & \cellcolor[HTML]{E7E6E6}\textbf{5.11} & \cellcolor[HTML]{E7E6E6}\textbf{7.11}  \\ \hline
                            & Atom                                   & 10.43                                 & 7.99                                  & 6.69                                  & 12.27                                 & 8.68                                  & 33.49                                  \\
                            & \cellcolor[HTML]{E7E6E6}Atom\_BiSup   & \cellcolor[HTML]{E7E6E6}\textbf{8.45} & \cellcolor[HTML]{E7E6E6}\textbf{7.18} & \cellcolor[HTML]{E7E6E6}\textbf{6.25} & \cellcolor[HTML]{E7E6E6}\textbf{8.76} & \cellcolor[HTML]{E7E6E6}\textbf{7.17} & \cellcolor[HTML]{E7E6E6}\textbf{18.68} \\
                            & QuaRot                                 & 9.89                                  & 7.18                                  & 6.26                                  & 13.92                                 & 8.79                                  & 39.95                                  \\
\multirow{-4}{*}{W3A3-g128} & \cellcolor[HTML]{E7E6E6}QuaRot\_BiSup & \cellcolor[HTML]{E7E6E6}\textbf{7.48} & \cellcolor[HTML]{E7E6E6}\textbf{6.40} & \cellcolor[HTML]{E7E6E6}\textbf{5.48} & \cellcolor[HTML]{E7E6E6}\textbf{7.74} & \cellcolor[HTML]{E7E6E6}\textbf{6.34} & \cellcolor[HTML]{E7E6E6}\textbf{13.69} \\
\hline
\end{tabular}
\end{table}

\begin{table}[ht]
\centering
\caption{Zero-shot accuracy results of Llama3-8B on Arc-Challenge (A-c), Arc-Easy (A-e), BoolQ (BQ), HellaSwag (HS), PIQA (PQ) and WinoGrande (WG). More results of other models are in \textsection \ref{apx:additional overall results}.}
\label{tab:llama3 acc}
\begin{tabular}{ccccccccc}
\hline
\multicolumn{2}{c}{\textbf{Llama3-8B / Acc$\uparrow$}}                      & \textbf{A-c}                 & \textbf{A-e}                      & \textbf{BQ}                         & \textbf{HS}                     & \textbf{PQ}                          & \textbf{WG}                    & \textbf{Avg.}                          \\ \hline
FP16                        & -                                      & 53.24                                  & 80.01                                  & 80.98                                  & 79.11                                  & 80.58                                  & 73.01                                  & 74.49                                  \\ \hline
                            & Atom                                   & 26.11                                  & 44.70                                  & 50.64                                  & 42.47                                  & 60.12                                  & 52.09                                  & 46.02                                  \\
                            & \cellcolor[HTML]{E7E6E6}Atom\_BiSup   & \cellcolor[HTML]{E7E6E6}\textbf{32.85} & \cellcolor[HTML]{E7E6E6}\textbf{63.76} & \cellcolor[HTML]{E7E6E6}\textbf{70.37} & \cellcolor[HTML]{E7E6E6}\textbf{57.61} & \cellcolor[HTML]{E7E6E6}\textbf{69.42} & \cellcolor[HTML]{E7E6E6}\textbf{59.83} & \cellcolor[HTML]{E7E6E6}\textbf{58.97} \\
                            & QuaRot                                 & 44.45                                  & 71.89                                  & 74.01                                  & 73.09                                  & 75.73                                  & 66.22                                  & 67.57                                  \\
\multirow{-4}{*}{W4A4}   & \cellcolor[HTML]{E7E6E6}QuaRot\_BiSup & \cellcolor[HTML]{E7E6E6}\textbf{48.55} & \cellcolor[HTML]{E7E6E6}\textbf{78.70} & \cellcolor[HTML]{E7E6E6}\textbf{77.43} & \cellcolor[HTML]{E7E6E6}\textbf{75.09} & \cellcolor[HTML]{E7E6E6}\textbf{77.04} & \cellcolor[HTML]{E7E6E6}\textbf{69.46} & \cellcolor[HTML]{E7E6E6}\textbf{71.05} \\ \hline
                            & Atom                                   & 48.21                                  & 76.89                                  & 74.43                                  & \textbf{75.27}                         & 77.42                                  & \textbf{67.56}                         & 69.96                                  \\
                            & \cellcolor[HTML]{E7E6E6}Atom\_BiSup   & \cellcolor[HTML]{E7E6E6}\textbf{50.94} & \cellcolor[HTML]{E7E6E6}\textbf{79.50} & \cellcolor[HTML]{E7E6E6}\textbf{79.72} & \cellcolor[HTML]{E7E6E6}74.08          & \cellcolor[HTML]{E7E6E6}\textbf{78.07} & \cellcolor[HTML]{E7E6E6}67.25          & \cellcolor[HTML]{E7E6E6}\textbf{71.59} \\
                            & QuaRot                                 & 47.78                                  & 75.55                                  & \textbf{79.63}                         & 76.17                                  & 76.93                                  & \textbf{71.11}                         & 71.20                                  \\
\multirow{-4}{*}{W4A4-g128} & \cellcolor[HTML]{E7E6E6}QuaRot\_BiSup & \cellcolor[HTML]{E7E6E6}\textbf{50.94} & \cellcolor[HTML]{E7E6E6}\textbf{79.17} & \cellcolor[HTML]{E7E6E6}78.96          & \cellcolor[HTML]{E7E6E6}\textbf{76.42} & \cellcolor[HTML]{E7E6E6}\textbf{78.73} & \cellcolor[HTML]{E7E6E6}70.96          & \cellcolor[HTML]{E7E6E6}\textbf{72.53} \\ \hline
                            & Atom                                   & 24.23                                  & 42.05                                  & \textbf{59.60}                         & 42.91                                  & 59.03                                  & 53.75                                  & 46.93                                  \\
                            & \cellcolor[HTML]{E7E6E6}Atom\_BiSup   & \cellcolor[HTML]{E7E6E6}\textbf{35.84} & \cellcolor[HTML]{E7E6E6}\textbf{64.39} & \cellcolor[HTML]{E7E6E6}55.69          & \cellcolor[HTML]{E7E6E6}\textbf{53.40} & \cellcolor[HTML]{E7E6E6}\textbf{68.72} & \cellcolor[HTML]{E7E6E6}\textbf{55.96} & \cellcolor[HTML]{E7E6E6}\textbf{55.67} \\
                            & QuaRot                                 & 23.81                                  & 39.90                                  & 56.76                                  & 39.95                                  & 57.83                                  & 51.22                                  & 44.91                                  \\
\multirow{-4}{*}{W3A3-g128} & \cellcolor[HTML]{E7E6E6}QuaRot\_BiSup & \cellcolor[HTML]{E7E6E6}\textbf{35.92} & \cellcolor[HTML]{E7E6E6}\textbf{67.63} & \cellcolor[HTML]{E7E6E6}\textbf{66.79} & \cellcolor[HTML]{E7E6E6}\textbf{61.21} & \cellcolor[HTML]{E7E6E6}\textbf{71.71} & \cellcolor[HTML]{E7E6E6}\textbf{59.91} & \cellcolor[HTML]{E7E6E6}\textbf{60.53} \\
\hline
\end{tabular}
\end{table}


To address RQ1, we evaluate the quantized models using language generation tasks and zero-shot tasks. The partial results of Llama family can be found in Table \ref{tab:llama wiki ppl} and Table \ref{tab:llama3 acc}, while the additional results are reported in Appendix (\textsection \ref{apx:additional overall results}). As demonstrated in the tables, BiSup consistently achieves performance improvements under the W4A4 and W3A3-g128 configurations, whereas its effect is less significant under W4A4-g128. This indicates that the benefits of BiSup are related to the quantization difficulty (i.e., the magnitude of quantization error) of different quantization configurations. Under more challenging quantization configurations (namely, W4A4 and W3A3-g128), BiSup can reliably deliver substantial benefits. However, when employing a simpler quantization configuration (i.e., W4A4-g128), BiSup does not show a marked advantage compared to the GPTQ algorithm used in original methods, which quantizes the tensor channel-by-channel and compensates for quantization errors in the unquantized channels. Note that BiSup utilizes the RTN algorithm as an alternative, which is more feasible for fine-tuning.
Refer to \textsection \ref{sec:error suppression analysis} for possible causes of suboptimal performance.

\subsection{Ablation Studies}
\label{sec:ablation studies}

\begin{table}[ht]
\centering
\caption{Ablation experiment results on different techniques used in BiSup. We report the WikiText2 perplexity of Llama3-8B (serving as a representative model) under three configurations.}
\label{tab:llama3 ablation}
\begin{tabular}{lccc}
\hline
\textbf{Llama-3-8B / PPL$\downarrow$} & \textbf{W4A4}    & \textbf{W4A4-g128} & \textbf{W3A3-g128} \\ \hline
Atom (GPTQ)                                        & 43.60            & 7.45               & 33.49              \\ \hline
Atom (RTN)                                         & 50.19            & 7.78               & 223.76             \\
+ Fine-Grained   Weight-Activation Clipping      & 25.39            & 8.21               & 50.29              \\
+   Soft-Constrained Weight-Activation Smoothing & 20.03            & 7.77               & 34.91              \\
+ Stabilized   Low-Rank Error Compensation      & 17.07            & 7.57               & 19.68              \\
+ Prompt   Mixed-Precision Quantization          & 16.32            & 7.55               & 18.68              \\ \hline
QuaRot (GPTQ)                                      & 8.22             & 7.33               & 39.95              \\ \hline
QuaRot (RTN)                                       & 19.08            & 9.28               & 4138.60            \\
+ Fine-Grained   Weight-Activation Clipping      & 10.03            & 8.12               & 174.81             \\
+   Soft-Constrained Weight-Activation Smoothing & 8.13             & 7.41               & 21.43              \\
+ Stabilized   Low-Rank Error Compensation      & 7.70             & 7.22               & 15.71              \\
+ Prompt   Mixed-Precision Quantization          & 7.42             & 7.11               & 13.69              \\
\hline
\end{tabular}
\end{table}

To validate the effectiveness of each technique used in BiSup (RQ2), we conduct ablation experiments starting from the original method (replacing GPTQ with RTN) and progressively incorporating the proposed  techniques. As illustrated in Table \ref{tab:llama3 ablation}, the four quantization techniques proposed in BiSup are effective across different quantization configurations and original methods. Except for the introduction of the Fine-Grained Weight-Activation Clipping mechanism on Atom (RTN) under W4A4-g128 (7.78 vs. 8.21). To this end, we directly introduce the Soft-Constrained Weight-Activation Smoothing mechanism on the top of Atom (RTN), but obtain similarly poor result (7.84 vs. 7.78). However, when both techniques were applied simultaneously, we achieve better result (7.77 vs. 7.78). Accordingly, we consider this anomaly to be random noise about the hyperparameters. Moreover, the first three proposed techniques have their corresponding original versions. In order to verify the effectiveness of the modifications in this paper, we perform additional comparative experiments in \textsection \ref{apx:comparative experiments}.

\subsection{Error Suppression Analysis}
\label{sec:error suppression analysis}
\begin{figure}[ht]
  \centering
  \includegraphics[width=\textwidth]{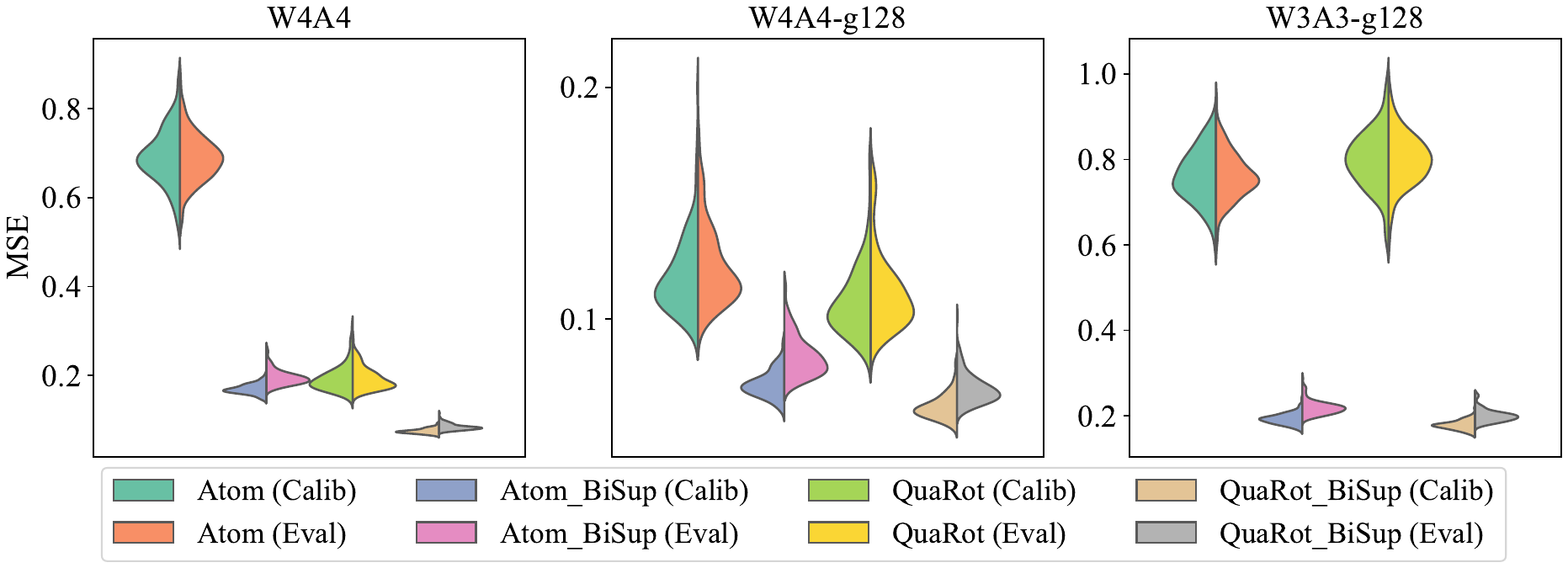}
  \caption{The mean square error (MSE$\downarrow$) of the activation of the last decoder layer in Llama3-8B. The dataset used for visualization is WikiText2, where Calib is sampled from the training set (and used to calibrate the quanitized model) and Eval is sampled from the test set. Explanation of notation: Atom (Calib) denotes the result of the Atom method on the Calib dataset of WikiText2. Note that due to algorithmic differences, Atom (or Atom\_BiSup) and QuaRot (or QuaRot\_BiSup) do not have the same activation, so MSE comparisons between these two types of methods make no sense. }
  \label{fig:mse}
\end{figure}
To tackle RQ2, we analyze the error propagation of different quantization methods in Llama3-8B. As depicted in Figure \ref{fig:mse}, the proposed BiSup can learn how to suppress error propagation on the calibration dataset and effectively generalize to other datasets. Corresponding to \textsection \ref{sec:overall results}, it can be observed that under W4A4 and W3A3-g128, BiSup significantly suppresses the quantization errors (the error suppression rate is about 70\% in the last layer). In contrast, under W4A4-g128, the error suppression rate is only about 30\%. One reason is that under W4A4-g128, the errors introduced by the quantized model are rather small, making it more difficult to optimize. Furthermore, we observe an inconsistency between quantization errors and final results: while in most cases smaller quantization errors typically correspond to lower perplexity, we note that under W4A4-g128, Atom\_BiSup 
(Eval) exhibits smaller quantization errors compared to Atom (Eval), but Atom (Eval) demonstrates lower perplexity (7.45 vs. 7.55 in Table \ref{tab:llama wiki ppl}). This finding indicates a difference between the currently widely used layer-wise optimization strategy based on mean square error and the overall optimization strategy based on cross-entropy loss of the final results. The layer-wise optimization strategy can effectively indicate the direction of optimization under challenging quantization configurations. However, when under simpler quantization configurations, it tends to fall into local optima. This finding reveals a potential reason for the suboptimal performance of BiSup under W4A4-g128. Besides, it is well known that the performance of gradient-based algorithms is related to the hyperparameters. Thus, we conduct experiments on the core hyperparameters of BiSup in \textsection \ref{apx:hyperparameter studies} for further discussion.

\subsection{Limitations and Improvements}
\label{sec:limitations and improvements}



In summary, BiSup is capable of improving performance under different settings, but it also shows some limitations and opportunities for improvement (RQ3).

\paragraph{Improvements of Layer-wise Optimization Strategy} As analyzed in \textsection \ref{sec:error suppression analysis}, the layer-wise optimization strategy, which does not adequately correlate with the final results, can lead to local optima. As an alternative, cross-layer optimization strategies that optimize multiple layers simultaneously could be considered. Additionally, different elements in the weights and activations have different importance (e.g. outliers are particularly important), which could be take into account when designing the loss function (e.g., by introducing the weighted mean square error).

\paragraph{Best Practices for Quantization-Aware Parameter-Efficient Fine-Tuning} As shown in the hyperparameter studies (\textsection \ref{apx:hyperparameter studies}), fine-tuning with more samples and iterations can yield better results. However, quantization techniques require a trade-off between memory consumption, time cost, and final performance. To this end, exploring the use of a small amount of high-quality data instead of large volumes of low-quality data, and investigating parameter initialization methods that align more closely with the quantization settings for faster convergence, could be beneficial.

\paragraph{Adequate Exploration of Prompt Mixed-Precision Quantization} As indicated by the ablation studies (\textsection \ref{sec:ablation studies}), while the prompt mixed-precision quantization strategy is effective, it has a limited impact on error suppression, with most of the contribution coming from fine-tuning. This is partly because the system prompts involved in our experiments are very short (e.g., non-instruction tuned models include only a single <bos> token). Experiments on more complex real-world scenarios are needed. Furthermore, the proposed prompt mixed-precision quantization strategy mainly focuses on system prompts. In practice, important tokens also exist in user prompts, which can be dynamically identified and preserved at high precision during inference \citep{yang2024no}.


\section{Conclusion}

Existing weight-activation quantization methods highlight optimizing the results of single matrix multiplication, overlooking the bidirectional propagation of quantization errors within the model, which includes the vertical accumulation within the same token and the horizontal diffusion across different tokens. To mitigate error propagation, We propose BiSup that constructs appropriate parameter spaces and applies quantization-aware parameter-efficient fine-tuning to compensate for quantization error in a timely manner, incorporating prompt mixed-precision quantization for the protection of important token interaction patterns. Extensive experiments on Llama and Qwen families validate the effectiveness of BiSup, and we conclude with a discussion of the limitations and improvements of BiSup.





\clearpage



\clearpage

\appendix

\setcounter{table}{0}
\setcounter{figure}{0}
\setcounter{algorithm}{0}

\renewcommand{\thetable}{A\arabic{table}}
\renewcommand{\thefigure}{A\arabic{figure}}
\renewcommand{\thealgorithm}{A\arabic{algorithm}}

\section{Appendix}
In this appendix, we provide more details as follows:
\begin{itemize}
\item \textsection \ref{apx:overall algorithm}: Pseudo-code for our BiSup algorithm.
\item \textsection \ref{apx:additional overall results}: Additional overall results for Llama and Qwen families.
\item \textsection \ref{apx:comparative experiments}: Comparative experiments between the proposed techniques and their original versions.
\item \textsection \ref{apx:hyperparameter studies}: Hyperparameter studies in quantization-aware parameter-efficient fine-tuning.
\end{itemize}

\subsection{Overall Algorithm}
\label{apx:overall algorithm}

The workflow of our BiSup is demonstrated in Algorithm \ref{alg:bisup}, which consists of four main steps: preprocessing the model according to the needs of Atom or QuaRot (Line 1), calculating and storing high-precision system prompts (Line 2), initializing learnable parameters and performing gradient optimization (Lines 3-16), and finally quantizing the model (Line 17).

\renewcommand{\algorithmicrequire}{\textbf{Input:}}
\renewcommand{\algorithmicensure}{\textbf{Output:}}

\begin{algorithm}
\caption{Overall Algorithm of BiSup}
\label{alg:bisup}
\begin{algorithmic}[1]
    \Require calibration dataset $X$, LLM $\mathcal{M}$, hadamard matrix $H$
    \Ensure quantized $\mathcal{\hat M}$
    \State reorder (or rotate) $\mathcal{M}$ using $X$ (or $H$) \Comment{preprocessing of Atom (or QuaRot)}
    \State calculate full-precision kv cache of system prompts with $\mathcal{M}$ \Comment{\textsection \ref{sec:prompt}}
    \State $X_{fp}=X_{int}=X$  \Comment{init the inputs of the transformer layers}
    \For{$\mathcal{L}_i$ in $\mathcal{M}$}: \Comment{layer-wise optimization}
        \State convert $\mathcal{L}_i$ from FP16 to BF16 \Comment{prevent numerical overflow}
        \State $X_{fp}=\mathcal{L}_i(X_{fp})$ \Comment{update full-precision inputs}
        \State init learnable parameters $\Theta=\{\Theta_1, \Theta_2, \Theta_3\}$ \Comment{\textsection \ref{sec:clipping}, \textsection \ref{sec:smoothing}, and \textsection \ref{sec:lorc}}
        \For{e in epochs}:
            \For{$(x_{fp}, x_{int})$ in $(X_{fp}, X_{int})$}: \Comment{\textsection \ref{sec:peft}}
                \State $x_{int} = \mathcal{L}_i(x_{int};\Theta)$ \Comment{quantize and forward}
                \State loss = ${\|x_{int} - x_{fp}\|}^2$ \Comment{MSELoss}
                \State backward and update $\Theta$ through gradient \Comment{with AdamW optimizer}
            \EndFor
        \EndFor
        \State $X_{int} = \mathcal{L}_i(X_{int};\Theta)$ \Comment{update quantized inputs}
        \State revert $\mathcal{L}_i$ from BF16 to FP16 \Comment{revert the data type}
        \State $\mathcal{\hat L}_i$ in $\mathcal{\hat M} \gets \mathrm{Quantize}(\mathcal{L}_i, \Theta)$ \Comment{quantize the transformer layer}
    \EndFor
\end{algorithmic}
\end{algorithm}

\subsection{Additional Overall Results}
\label{apx:additional overall results}

In this section, we provide additional overall results for Llama and Qwen families. The list of tables is as follows:
\begin{itemize}
\item Table \ref{tab:llama c4 ppl}: C4 perplexity results of Llama family.
\item Table \ref{tab:llama w4a4 acc}: Zero-shot accuracy results of Llama family under W4A4.
\item Table \ref{tab:llama w4a4-g128 acc}: Zero-shot accuracy results of Llama family under W4A4-g128.
\item Table \ref{tab:llama w3a3-g128 acc}: Zero-shot accuracy results of Llama family under W3A3-g128.
\item Table \ref{tab:llama3-instruct acc}: Zero-shot accuracy results of Llama3-8B-Instruct.
\item Table \ref{tab:qwen wiki ppl}: WikiText2 perplexity results of Qwen family.
\item Table \ref{tab:qwen c4 ppl}: C4 perplexity results of Qwen family.
\item Table \ref{tab:qwen w4a4 acc}: Zero-shot accuracy results of Qwen family under W4A4.
\item Table \ref{tab:qwen w4a4-g128 acc}: Zero-shot accuracy results of Qwen family under W4A4-g128.
\item Table \ref{tab:qwen w3a3-g128 acc}: Zero-shot accuracy results of Qwen family under W3A3-g128.
\end{itemize}

\begin{table}[ht]
\centering
\caption{C4 perplexity results of Llama family.}
\label{tab:llama c4 ppl}
\begin{tabular}{cccccccc}
\hline
\multicolumn{2}{c}{\textbf{Llama1\&2\&3   /  PPL$\downarrow$}}                & \textbf{1-7B}                          & \textbf{1-13B}                        & \textbf{1-30B}                        & \textbf{2-7B}                          & \textbf{2-13B}                        & \textbf{3-8B}                          \\ \hline
FP16                        & -                                      & 7.08                                   & 6.60                                  & 5.98                                  & 6.97                                   & 6.46                                  & 8.88                                   \\ \hline
                            & Atom                                   & 11.97                                  & 11.02                                 & 9.83                                  & 13.07                                  & 11.12                                 & 58.34                                  \\
                            & \cellcolor[HTML]{E7E6E6}Atom\_BiSup   & \cellcolor[HTML]{E7E6E6}\textbf{10.23} & \cellcolor[HTML]{E7E6E6}\textbf{9.31} & \cellcolor[HTML]{E7E6E6}\textbf{8.46} & \cellcolor[HTML]{E7E6E6}\textbf{10.76} & \cellcolor[HTML]{E7E6E6}\textbf{9.77} & \cellcolor[HTML]{E7E6E6}\textbf{22.30} \\
                            & QuaRot                                 & 7.81                                   & 7.06                                  & 6.37                                  & 7.80                                   & 7.07                                  & 11.87                                  \\
\multirow{-4}{*}{W4A4}   & \cellcolor[HTML]{E7E6E6}QuaRot\_BiSup & \cellcolor[HTML]{E7E6E6}\textbf{7.55}  & \cellcolor[HTML]{E7E6E6}\textbf{6.94} & \cellcolor[HTML]{E7E6E6}\textbf{6.27} & \cellcolor[HTML]{E7E6E6}\textbf{7.57}  & \cellcolor[HTML]{E7E6E6}\textbf{6.84} & \cellcolor[HTML]{E7E6E6}\textbf{10.88} \\
\hline
\end{tabular}
\end{table}

\begin{table}[ht]
\centering
\caption{Zero-shot accuracy results of Llama family under W4A4.}
\label{tab:llama w4a4 acc}
\begin{tabular}{ccccccccc}
\hline
\multicolumn{2}{c}{\textbf{Llama1\&2   / Acc$\uparrow$}}                           & \textbf{A-c}       & \textbf{A-e}                           & \textbf{BQ}                            & \textbf{HS}                            & \textbf{PQ}                            & \textbf{WG}                            & \textbf{Avg.}                          \\ \hline
                                       & FP16                                   & 44.20                                  & 75.46                                  & 76.79                                  & 76.05                                  & 79.27                                  & 70.24                                  & 70.33                                  \\ \cline{2-9}
                                       & Atom                                   & \textbf{37.20}                         & 62.29                                  & 65.29                                  & 63.94                                  & 70.13                                  & 58.56                                  & 59.57                                  \\
                                       & \cellcolor[HTML]{E7E6E6}Atom\_BiSup   & \cellcolor[HTML]{E7E6E6}34.98          & \cellcolor[HTML]{E7E6E6}\textbf{66.92} & \cellcolor[HTML]{E7E6E6}\textbf{66.70} & \cellcolor[HTML]{E7E6E6}\textbf{64.11} & \cellcolor[HTML]{E7E6E6}\textbf{73.34} & \cellcolor[HTML]{E7E6E6}\textbf{62.35} & \cellcolor[HTML]{E7E6E6}\textbf{61.40} \\
                                       & QuaRot                                 & 40.10                                  & 72.26                                  & 75.38                                  & \textbf{72.87}                         & 76.88                                  & 67.48                                  & 67.50                                  \\
\hline
\end{tabular}
\end{table}

\begin{table}[ht]
\centering
\caption{Zero-shot accuracy results of Llama family under W4A4-g128.}
\label{tab:llama w4a4-g128 acc}
\begin{tabular}{ccccccccc}
\hline
\multicolumn{2}{c}{\textbf{Llama1\&2   / Acc$\uparrow$}}                      & \textbf{A-c}       & \textbf{A-e}                           & \textbf{BQ}                            & \textbf{HS}                            & \textbf{PQ}                            & \textbf{WG}                            & \textbf{Avg.}                          \\ \hline
                                       & FP16                                   & 44.20                                  & 75.46                                  & 76.79                                  & 76.05                                  & 79.27                                  & 70.24                                  & 70.33                                  \\ \cline{2-9}
                                       & Atom                                   & 40.87                                  & 72.60                                  & \textbf{74.13}                         & \textbf{72.66}                         & \textbf{76.66}                         & \textbf{66.93}                         & \textbf{67.31}                         \\
                                       & \cellcolor[HTML]{E7E6E6}Atom\_BiSup   & \cellcolor[HTML]{E7E6E6}\textbf{41.21} & \cellcolor[HTML]{E7E6E6}\textbf{72.98} & \cellcolor[HTML]{E7E6E6}73.39          & \cellcolor[HTML]{E7E6E6}72.17          & \cellcolor[HTML]{E7E6E6}76.61          & \cellcolor[HTML]{E7E6E6}66.61          & \cellcolor[HTML]{E7E6E6}67.16          \\
                                       & QuaRot                                 & \textbf{43.34}                         & 72.98                                  & \textbf{75.50}                         & 73.16                                  & 77.86                                  & \textbf{68.82}                         & \textbf{68.61}                         \\
\hline
\end{tabular}
\end{table}

\begin{table}[ht]
\centering
\caption{Zero-shot accuracy results of Llama family under W3A3-g128.}
\label{tab:llama w3a3-g128 acc}
\begin{tabular}{ccccccccc}
\hline
\multicolumn{2}{c}{\textbf{Llama1\&2   /  Acc$\uparrow$}}                      & \textbf{A-c}       & \textbf{A-e}                           & \textbf{BQ}                            & \textbf{HS}                            & \textbf{PQ}                            & \textbf{WG}                            & \textbf{Avg.}                          \\ \hline
                                       & FP16                                   & 44.20                                  & 75.46                                  & 76.79                                  & 76.05                                  & 79.27                                  & 70.24                                  & 70.33                                  \\ \cline{2-9}
                                       & Atom                                   & 29.69                                  & 59.22                                  & 63.85                                  & 56.40                                  & 68.12                                  & 56.04                                  & 55.55                                  \\
                                       & \cellcolor[HTML]{E7E6E6}Atom\_BiSup   & \cellcolor[HTML]{E7E6E6}\textbf{33.62} & \cellcolor[HTML]{E7E6E6}\textbf{64.98} & \cellcolor[HTML]{E7E6E6}\textbf{66.18} & \cellcolor[HTML]{E7E6E6}\textbf{59.44} & \cellcolor[HTML]{E7E6E6}\textbf{70.78} & \cellcolor[HTML]{E7E6E6}\textbf{60.38} & \cellcolor[HTML]{E7E6E6}\textbf{59.23} \\
                                       & QuaRot                                 & 31.40                                  & 59.64                                  & 64.13                                  & 56.91                                  & 68.66                                  & 56.12                                  & 56.14                                  \\
\hline
\end{tabular}
\end{table}

\begin{table}[ht]
\centering
\caption{Zero-shot accuracy results of Llama3-8B-Instruct.}
\label{tab:llama3-instruct acc}
\begin{tabular}{ccccccccc}
\hline
\multicolumn{2}{c}{\textbf{Llama3-8B-Instruct   /  Acc$\uparrow$}}             & \textbf{A-c}                           & \textbf{A-e}                           & \textbf{BQ}                            & \textbf{HS}                            & \textbf{PQ}                            & \textbf{WG}                            & \textbf{Avg.}                          \\ \hline
FP16                        & -                                      & 57.08                                  & 81.52                                  & 83.27                                  & 75.80                                  & 79.00                                  & 72.45                                  & 74.85                                  \\ \hline
                            & Atom                                   & 28.84                                  & 47.85                                  & 52.75                                  & 46.52                                  & 61.15                                  & 53.99                                  & 48.52                                  \\
                            & \cellcolor[HTML]{E7E6E6}Atom\_BiSup   & \cellcolor[HTML]{E7E6E6}\textbf{36.35} & \cellcolor[HTML]{E7E6E6}\textbf{66.62} & \cellcolor[HTML]{E7E6E6}\textbf{76.12} & \cellcolor[HTML]{E7E6E6}\textbf{56.25} & \cellcolor[HTML]{E7E6E6}\textbf{67.19} & \cellcolor[HTML]{E7E6E6}\textbf{60.38} & \cellcolor[HTML]{E7E6E6}\textbf{60.48} \\
                            & QuaRot                                 & 46.25                                  & 75.93                                  & 79.85                                  & 70.98                                  & 75.57                                  & 66.77                                  & 69.22                                  \\
\hline
\end{tabular}
\end{table}

\begin{table}[ht]
\centering
\caption{WikiText2 perplexity results of Qwen family.}
\label{tab:qwen wiki ppl}
\begin{tabular}{cccccccc}
\hline
\multicolumn{2}{c}{\textbf{Qwen1.5 / PPL$\downarrow$}}           & \textbf{0.5B}                          & \textbf{1.8B}                          & \textbf{4B}                            & \textbf{7B}                            & \textbf{14B}                           & \textbf{32B}                          \\ \hline
FP16                        & -                                    & 14.81                                  & 11.45                                  & 8.94                                   & 7.95                                   & 7.45                                   & 6.17                                  \\ \hline
                            & Atom                                 & 34.47                                  & 20.20                                  & 15.04                                  & 14.69                                  & 14.67                                  & 16.65                                 \\
\multirow{-2}{*}{W4A4}      & \cellcolor[HTML]{E7E6E6}Atom\_BiSup & \cellcolor[HTML]{E7E6E6}\textbf{21.75} & \cellcolor[HTML]{E7E6E6}\textbf{15.76} & \cellcolor[HTML]{E7E6E6}\textbf{14.21} & \cellcolor[HTML]{E7E6E6}\textbf{11.92} & \cellcolor[HTML]{E7E6E6}\textbf{10.17} & \cellcolor[HTML]{E7E6E6}\textbf{8.65} \\ \hline
                            & Atom                                 & 22.40                                  & 13.31                                  & \textbf{10.00}                         & \textbf{8.72}                          & \textbf{7.95}                          & 6.58                                  \\
\multirow{-2}{*}{W4A4-g128} & \cellcolor[HTML]{E7E6E6}Atom\_BiSup & \cellcolor[HTML]{E7E6E6}\textbf{18.07} & \cellcolor[HTML]{E7E6E6}\textbf{12.99} & \cellcolor[HTML]{E7E6E6}10.69          & \cellcolor[HTML]{E7E6E6}8.97           & \cellcolor[HTML]{E7E6E6}7.98           & \cellcolor[HTML]{E7E6E6}\textbf{6.57} \\
\hline
\end{tabular}
\end{table}

\begin{table}[ht]
\centering
\caption{C4 perplexity results of Qwen family.}
\label{tab:qwen c4 ppl}
\begin{tabular}{cccccccc}
\hline
\multicolumn{2}{c}{\textbf{Qwen1.5 / PPL$\downarrow$}}                   & \textbf{0.5B}                          & \textbf{1.8B}                          & \textbf{4B}                            & \textbf{7B}                            & \textbf{14B}                           & \textbf{32B}                           \\ \hline
FP16                        & -                                    & 17.38                                  & 13.96                                  & 12.12                                  & 11.01                                  & 10.22                                  & 8.34                                   \\ \hline
                            & Atom                                 & 43.99                                  & 25.01                                  & 20.20                                  & 20.06                                  & 20.83                                  & 23.08                                  \\
\multirow{-2}{*}{W4A4}      & \cellcolor[HTML]{E7E6E6}Atom\_BiSup & \cellcolor[HTML]{E7E6E6}\textbf{30.97} & \cellcolor[HTML]{E7E6E6}\textbf{21.58} & \cellcolor[HTML]{E7E6E6}\textbf{20.06} & \cellcolor[HTML]{E7E6E6}\textbf{17.31} & \cellcolor[HTML]{E7E6E6}\textbf{14.77} & \cellcolor[HTML]{E7E6E6}\textbf{12.46} \\ \hline
                            & Atom                                 & 26.68                                  & \textbf{16.45}                         & \textbf{13.62}                         & \textbf{12.15}                         & \textbf{10.95}                         & \textbf{8.94}                          \\
\multirow{-2}{*}{W4A4-g128} & \cellcolor[HTML]{E7E6E6}Atom\_BiSup & \cellcolor[HTML]{E7E6E6}\textbf{23.40} & \cellcolor[HTML]{E7E6E6}16.52          & \cellcolor[HTML]{E7E6E6}14.81          & \cellcolor[HTML]{E7E6E6}12.49          & \cellcolor[HTML]{E7E6E6}11.23          & \cellcolor[HTML]{E7E6E6}9.09           \\
\hline
\end{tabular}
\end{table}

\begin{table}[ht]
\centering
\caption{Zero-shot accuracy results of Qwen family under W4A4.}
\begin{tabular}{ccccccccc}
\hline
\multicolumn{2}{c}{\textbf{Qwen1.5 / Acc$\uparrow$}}           & \textbf{A-c}                           & \textbf{A-e}                           & \textbf{BQ}                            & \textbf{HS}                            & \textbf{PQ}                            & \textbf{WG}                            & \textbf{Avg.}                          \\ \hline
                       & FP16                                 & 29.18                                  & 57.11                                  & 49.66                                  & 49.28                                  & 69.48                                  & 55.25                                  & 51.66                                  \\ \cline{2-9}
                       & Atom                                 & 24.91                                  & 42.85                                  & 52.39                                  & 36.10                                  & 59.90                                  & 48.62                                  & 44.13                                  \\
\multirow{-3}{*}{0.5B} & \cellcolor[HTML]{E7E6E6}Atom\_BiSup & \cellcolor[HTML]{E7E6E6}\textbf{26.54} & \cellcolor[HTML]{E7E6E6}\textbf{47.73} & \cellcolor[HTML]{E7E6E6}\textbf{61.68} & \cellcolor[HTML]{E7E6E6}\textbf{38.89} & \cellcolor[HTML]{E7E6E6}\textbf{62.30} & \cellcolor[HTML]{E7E6E6}\textbf{51.62} & \cellcolor[HTML]{E7E6E6}\textbf{48.13} \\ \hline
                       & FP16                                 & 34.56                                  & 64.77                                  & 66.39                                  & 60.88                                  & 74.43                                  & 61.48                                  & 60.42                                  \\
\hline
\end{tabular}
\label{tab:qwen w4a4 acc}
\end{table}

\begin{table}[ht]
\centering
\caption{Zero-shot accuracy results of Qwen family under W4A4-g128.}
\label{tab:qwen w4a4-g128 acc}
\begin{tabular}{ccccccccc}
\hline
\multicolumn{2}{c}{\textbf{Qwen1.5 / Acc$\uparrow$}}      & \textbf{A-c}                           & \textbf{A-e}                           & \textbf{BQ}                            & \textbf{HS}                            & \textbf{PQ}                            & \textbf{WG}                            & \textbf{Avg.}                          \\ \hline
                       & FP16                                 & 29.18                                  & 57.11                                  & 49.66                                  & 49.28                                  & 69.48                                  & 55.25                                  & 51.66                                  \\ \cline{2-9}
                       & Atom                                 & \textbf{27.39}                         & \textbf{49.92}                         & 54.98                                  & 42.01                                  & 64.15                                  & 50.99                                  & 48.24                                  \\
\multirow{-3}{*}{0.5B} & \cellcolor[HTML]{E7E6E6}Atom\_BiSup & \cellcolor[HTML]{E7E6E6}27.30          & \cellcolor[HTML]{E7E6E6}49.20          & \cellcolor[HTML]{E7E6E6}\textbf{60.40} & \cellcolor[HTML]{E7E6E6}\textbf{42.69} & \cellcolor[HTML]{E7E6E6}\textbf{64.74} & \cellcolor[HTML]{E7E6E6}\textbf{53.43} & \cellcolor[HTML]{E7E6E6}\textbf{49.63} \\ \hline
                       & FP16                                 & 34.56                                  & 64.77                                  & 66.39                                  & 60.88                                  & 74.43                                  & 61.48                                  & 60.42                                  \\
\hline
\end{tabular}
\end{table}

\begin{table}[ht]
\centering
\caption{Zero-shot accuracy results of Qwen family under W3A3-g128.}
\label{tab:qwen w3a3-g128 acc}
\begin{tabular}{ccccccccc}
\hline
\multicolumn{2}{c}{\textbf{Qwen1.5 / Acc$\uparrow$}}      & \textbf{A-c}                           & \textbf{A-e}                           & \textbf{BQ}                            & \textbf{HS}                            & \textbf{PQ}                            & \textbf{WG}                            & \textbf{Avg.}                          \\ \hline
                       & FP16                                 & 29.18                                  & 57.11                                  & 49.66                                  & 49.28                                  & 69.48                                  & 55.25                                  & 51.66                                  \\ \cline{2-9}
                       & Atom                                 & \textbf{24.91}                         & 30.01                                  & 50.21                                  & 27.29                                  & 52.39                                  & \textbf{50.36}                         & 39.20                                  \\
\multirow{-3}{*}{0.5B} & \cellcolor[HTML]{E7E6E6}Atom\_BiSup & \cellcolor[HTML]{E7E6E6}22.35          & \cellcolor[HTML]{E7E6E6}\textbf{41.04} & \cellcolor[HTML]{E7E6E6}\textbf{61.28} & \cellcolor[HTML]{E7E6E6}\textbf{32.56} & \cellcolor[HTML]{E7E6E6}\textbf{58.16} & \cellcolor[HTML]{E7E6E6}49.96          & \cellcolor[HTML]{E7E6E6}\textbf{44.23} \\ \hline
                       & FP16                                 & 34.56                                  & 64.77                                  & 66.39                                  & 60.88                                  & 74.43                                  & 61.48                                  & 60.42                                  \\
\hline
\end{tabular}
\end{table}

\clearpage

\subsection{Comparative Experiments}
\label{apx:comparative experiments}

\begin{figure}[ht]
  \centering
  \includegraphics[width=\textwidth]{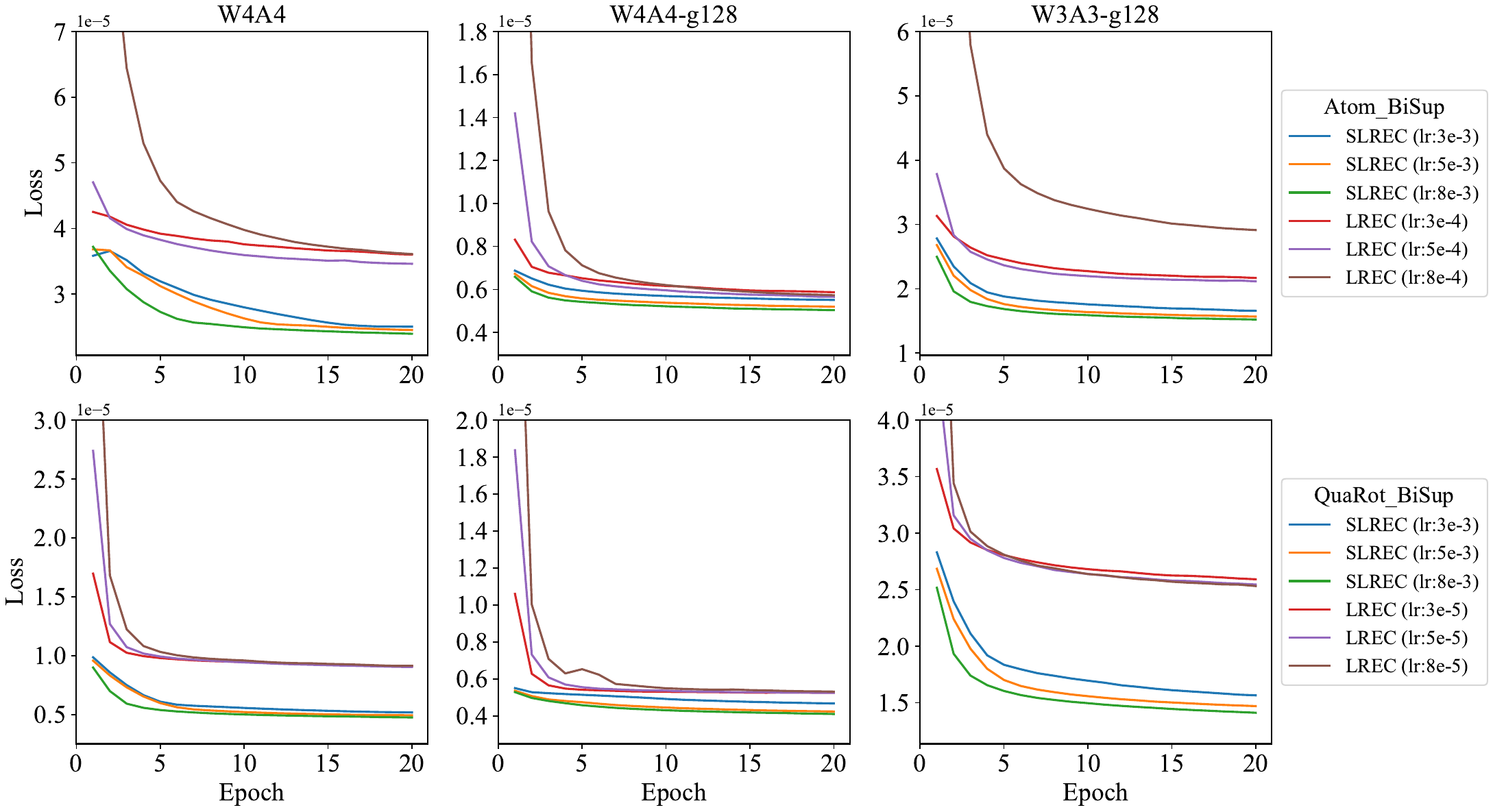}
  \caption{Loss curves on the first layer of Llama3-8B under different settings. (S)LREC denotes (Stabilized) Low-Rank Error Compensation.}
  \label{fig:llama convergence}
\end{figure}


\begin{figure}[ht]
  \centering
  \includegraphics[width=\textwidth]{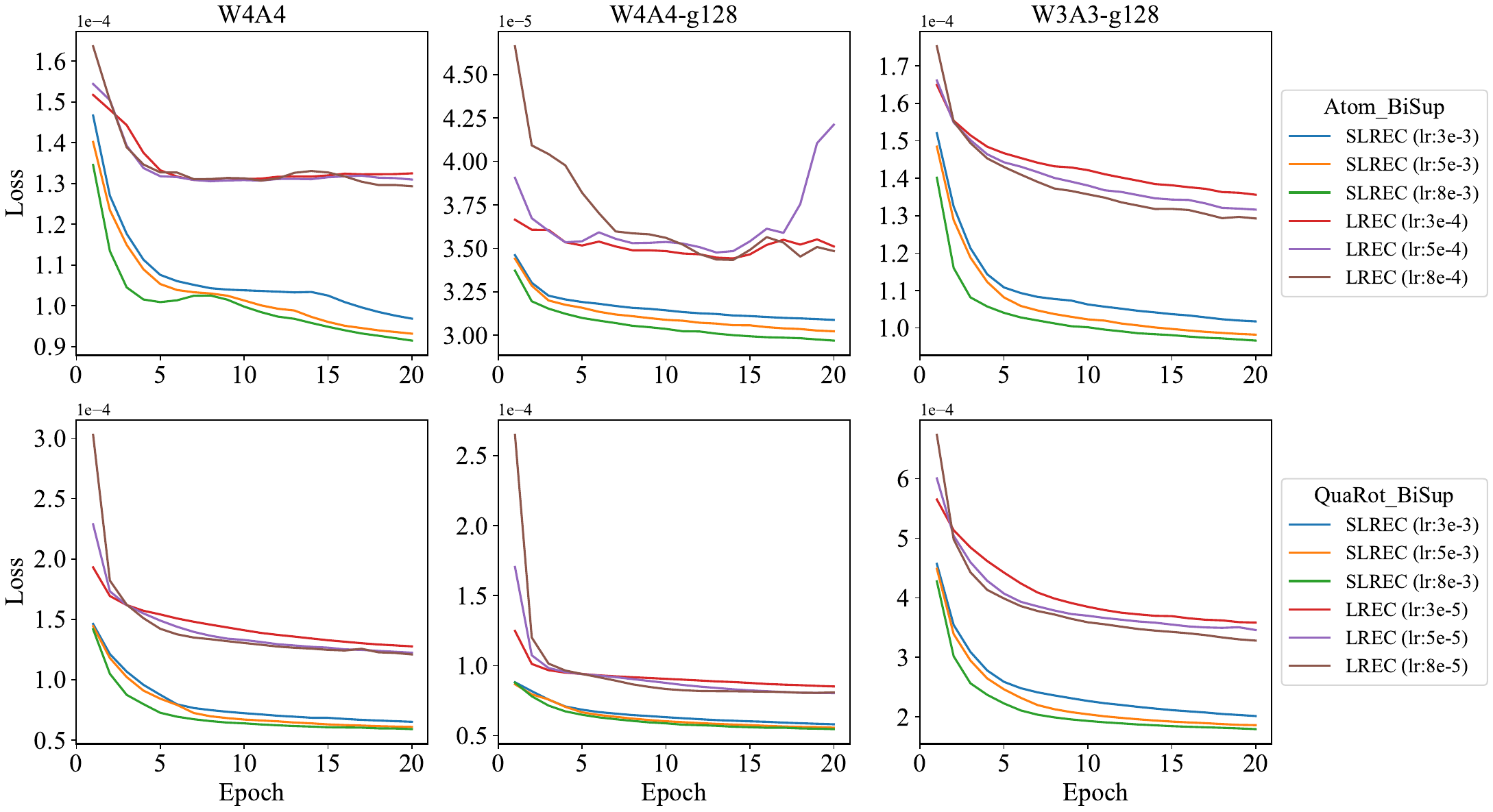}
  \caption{Loss curves on the first layer of Qwen1.5-7B under different settings. (S)LREC denotes (Stabilized) Low-Rank Error Compensation.}
  \label{fig:qwen convergence}
\end{figure}

\begin{table}[ht]
\centering
\caption{WikiText2 perplexity results of Llama-3-8B with different techniques. The notations are explained as follows: (F)WAC denotes (Fine-grained) Weight-Activation Clipping and (S)WAS means (Soft-constrained) Weight-Activation Smoothing.}
\label{tab:llama3 comparative experiments}
\begin{tabular}{lccc}
\hline
\textbf{Llama-3-8B / PPL$\downarrow$} & \textbf{W4A4}  & \textbf{W4A4-g128} & \textbf{W3A3-g128} \\ \hline
Atom (RTN)                             & 50.19          & 7.78               & 223.76             \\ \hline
Atom (RTN) + WAC                       & 25.39          & 8.44               & 83.40              \\
\rowcolor[HTML]{E7E6E6} 
Atom (RTN) + FWAC                      & \textbf{25.39} & \textbf{8.21}      & \textbf{50.29}     \\ \hline
Atom (RTN) + FWAC + WAS                & \textbf{15.21} & 7.80               & 34.94              \\
\rowcolor[HTML]{E7E6E6} 
Atom (RTN) + FWAC + SWAS               & 20.03          & \textbf{7.77}      & \textbf{34.91}     \\ \hline
QuaRot (RTN)                           & 19.08          & 9.28               & 4138.60            \\ \hline
QuaRot (RTN) + WAC                     & 10.03          & 8.82               & 218.19             \\
\rowcolor[HTML]{E7E6E6} 
QuaRot (RTN) + FWAC                    & \textbf{10.03} & \textbf{8.12}      & \textbf{174.81}    \\ \hline
QuaRot (RTN) + FWAC + WAS              & 8.15           & 7.56               & 204.91             \\
\rowcolor[HTML]{E7E6E6} 
QuaRot (RTN) + FWAC + SWAS             & \textbf{8.13}  & \textbf{7.41}      & \textbf{21.43}    \\  \hline
\end{tabular}
\end{table}

In order to verify the effectiveness of the modifications on the proposed techniques, we conducted comparative experiments with their original versions. Specifically, the Weight-Activation Clipping uses the same clip value for the entire tensor (\textsection \ref{sec:clipping}), the Weight-Activation Smoothing employs the same smoothing factor for weight and activation (\textsection \ref{sec:smoothing}), and the Low-Rank Error Compensation adopts the form similar to LoRA (\textsection \ref{sec:lorc}). The results of clipping and smoothing techniques are presented in Table \ref{tab:llama3 comparative experiments}. It can be seen that FWAC always outperforms WAC, while SWAS outperforms WAS in most cases, suggesting the validity of the modifications and that different combinations of techniques may be required for different settings. As depicted in Figures \ref{fig:llama convergence} and \ref{fig:qwen convergence}, compared to LREC, SLREC achieves better convergence and does not require choosing different learning rates for different settings to obtain stable performance, verifying the effectiveness of the proposed improvement.

\subsection{Hyperparameter Studies}
\label{apx:hyperparameter studies}

\begin{table}[ht]
\centering
\caption{WikiText2 perplexity results of Llama-3-8B with different \textbf{S}amples and \textbf{I}terations.}
\begin{tabular}{c|c|ccc|ccc}
\hline
                                                      &                    & \multicolumn{3}{c|}{\textbf{Atom\_BiSup}}                                                     & \multicolumn{3}{c}{\textbf{QuaRot\_BiSup}}                                                    \\ \cline{3-8} 
\multirow{-2}{*}{\textbf{Llama-3-8B / PPL$\downarrow$}} & \multirow{-2}{*}{\diagbox[dir=NW]{\textbf{S}}{\textbf{I}}} & \textbf{5}                    & \textbf{10}                   & \textbf{20}                   & \textbf{5}                    & \textbf{10}                   & \textbf{20}                   \\ \hline
                                                      & \textbf{64}        & \cellcolor[HTML]{5A8AC6}22.86 & \cellcolor[HTML]{DEE7F5}16.08 & \cellcolor[HTML]{FAC6C9}14.05 & \cellcolor[HTML]{5A8AC6}7.84  & \cellcolor[HTML]{DFE8F5}7.41  & \cellcolor[HTML]{FAFBFF}7.33  \\
                                                      & \textbf{128}       & \cellcolor[HTML]{DAE4F3}16.32 & \cellcolor[HTML]{FCFCFF}14.53 & \cellcolor[HTML]{F88587}13.46 & \cellcolor[HTML]{DDE6F4}7.42  & \cellcolor[HTML]{FAD0D3}7.30  & \cellcolor[HTML]{F8696B}7.27  \\
\multirow{-3}{*}{\textbf{W4A4}}                       & \textbf{256}       & \cellcolor[HTML]{FBFBFF}14.61 & \cellcolor[HTML]{F9AFB1}13.84 & \cellcolor[HTML]{F8696B}13.21 & \cellcolor[HTML]{FCFCFF}7.32  & \cellcolor[HTML]{FAC8CA}7.30  & \cellcolor[HTML]{F99FA2}7.28  \\ \hline
                                                      & \textbf{64}        & \cellcolor[HTML]{5A8AC6}7.62  & \cellcolor[HTML]{C7D7ED}7.53  & \cellcolor[HTML]{FBF9FC}7.48  & \cellcolor[HTML]{5A8AC6}7.13  & \cellcolor[HTML]{DAE4F3}7.09  & \cellcolor[HTML]{FCFCFF}7.08  \\
                                                      & \textbf{128}       & \cellcolor[HTML]{B0C7E5}7.55  & \cellcolor[HTML]{E0E8F5}7.50  & \cellcolor[HTML]{FBE9EC}7.47  & \cellcolor[HTML]{A3BDE0}7.11  & \cellcolor[HTML]{FACFD2}7.07  & \cellcolor[HTML]{FBEDF0}7.08  \\
\multirow{-3}{*}{\textbf{W4A4-g128}}                  & \textbf{256}       & \cellcolor[HTML]{FCFCFF}7.48  & \cellcolor[HTML]{F99193}7.44  & \cellcolor[HTML]{F8696B}7.42  & \cellcolor[HTML]{FBD7DA}7.07  & \cellcolor[HTML]{EBF0F9}7.08  & \cellcolor[HTML]{F8696B}7.06  \\ \hline
                                                      & \textbf{64}        & \cellcolor[HTML]{5A8AC6}22.79 & \cellcolor[HTML]{DCE6F4}17.37 & \cellcolor[HTML]{FBDEE1}15.72 & \cellcolor[HTML]{5A8AC6}16.18 & \cellcolor[HTML]{D4E0F1}13.37 & \cellcolor[HTML]{FBE2E5}12.28 \\
                                                      & \textbf{128}       & \cellcolor[HTML]{BDD0E9}18.68 & \cellcolor[HTML]{FCFCFF}16.03 & \cellcolor[HTML]{F99A9D}14.97 & \cellcolor[HTML]{C6D6EC}13.69 & \cellcolor[HTML]{FCFCFF}12.42 & \cellcolor[HTML]{F88B8D}11.82 \\
\multirow{-3}{*}{\textbf{W3A3-g128}}                  & \textbf{256}       & \cellcolor[HTML]{EFF3FB}16.58 & \cellcolor[HTML]{F9A3A5}15.07 & \cellcolor[HTML]{F8696B}14.43 & \cellcolor[HTML]{F2F5FC}12.65 & \cellcolor[HTML]{FABFC2}12.10 & \cellcolor[HTML]{F8696B}11.63 \\ \hline
\end{tabular}
\label{tab:hyperparameter:samples and iterations}
\end{table}

\begin{table}[ht]
\centering
\caption{WikiText2 perplexity results of Llama-3-8B with different Rank.}
\label{tab:hyperparameter:ranks}
\begin{tabular}{c|ccc|ccc}
\hline
                                                       & \multicolumn{3}{c|}{\textbf{Atom\_BiSup}}                                                     & \multicolumn{3}{c}{\textbf{QuaRot\_BiSup}}                                                    \\ \cline{2-7} 
\multirow{-2}{*}{\textbf{Llama-3-8B / PPL$\downarrow$}} & \textbf{16}                   & \textbf{32}                   & \textbf{64}                   & \textbf{16}                   & \textbf{32}                   & \textbf{64}                   \\ \hline
\textbf{W4A4}                                          & \cellcolor[HTML]{5A8AC6}16.81 & \cellcolor[HTML]{FCFCFF}16.32 & \cellcolor[HTML]{F8696B}16.07 & \cellcolor[HTML]{5A8AC6}7.44  & \cellcolor[HTML]{FCFCFF}7.42  & \cellcolor[HTML]{F8696B}7.39  \\
\textbf{W4A4-g128}                                     & \cellcolor[HTML]{5A8AC6}7.56  & \cellcolor[HTML]{FCFCFF}7.55  & \cellcolor[HTML]{F8696B}7.51  & \cellcolor[HTML]{FCFCFF}7.10  & \cellcolor[HTML]{5A8AC6}7.11  & \cellcolor[HTML]{F8696B}7.07  \\
\textbf{W3A3-g128}                                     & \cellcolor[HTML]{5A8AC6}19.38 & \cellcolor[HTML]{FCFCFF}18.68 & \cellcolor[HTML]{F8696B}17.88 & \cellcolor[HTML]{5A8AC6}13.91 & \cellcolor[HTML]{FCFCFF}13.69 & \cellcolor[HTML]{F8696B}13.33 \\ \hline
\end{tabular}
\end{table}

Experience has shown that algorithms based on gradient optimization are sensitive to the choice of hyperparameters. In this section, we conduct experiments on the main hyperparameters of BiSup, which are directly linked to the overhead of fine-tuning, including the number of training \textbf{samples}, the number of fine-tuning \textbf{iterations}, and the \textbf{rank} of the low-rank error compensation matrix. The results are shown in Tables \ref{tab:hyperparameter:samples and iterations} and \ref{tab:hyperparameter:ranks}. Generally speaking, BiSup can produce better results with more samples, larger iterations, and higher rank. However, the required memory footprint and computation time increase accordingly, as illustrated in Table \ref{tab:hyperparameter:cost}. In our experiments, we use 128 samples to fine-tune 5 iterations with a rank of 32. Obviously, the results produced in this configuration do not represent the optimal results for BiSup. Since our aim is to verify the general validity of BiSup, relatively simple configurations are chosen for a wide range of experiments, which can significantly reduce the cost of the experiments and have no impact on the main conclusions.

\clearpage

\begin{table}[ht]
\centering
\caption{The memory footprint and quantization time of Llama models in W4A4-g128.}
\label{tab:hyperparameter:cost}
\begin{tabular}{cc|ccc|ccc}
\hline
\multicolumn{2}{c|}{\multirow{2}{*}{\textbf{Samples - Iterations - Rank}}} & \multicolumn{3}{c|}{\textbf{Atom\_BiSup}}       & \multicolumn{3}{c}{\textbf{QuaRot\_BiSup}}      \\ \cline{3-8} 
\multicolumn{2}{c|}{}                                                    & \textbf{1-30B} & \textbf{2-13B} & \textbf{3-8B} & \textbf{1-30B} & \textbf{2-13B} & \textbf{3-8B} \\ \hline
\multicolumn{1}{c|}{\multirow{2}{*}{64 - 5 - 32}}      & Time (min)      & 74             & 32             & 17            & 71             & 32             & 17            \\
\multicolumn{1}{c|}{}                                  & Memory (MiB)    & 18,414         & 13,684         & 10,192        & 16,388         & 12,094         & 9,372         \\ \hline
\multicolumn{1}{c|}{\multirow{2}{*}{128 - 5 - 32}}     & Time (min)      & 148            & 62             & 34            & 141            & 59             & 34            \\
\multicolumn{1}{c|}{}                                  & Memory (MiB)    & 22,126         & 17,012         & 12,262        & 19,716         & 15,422         & 11,420        \\ \hline
\multicolumn{1}{c|}{\multirow{2}{*}{128 - 10 - 32}}    & Time (min)      & 285            & 120            & 66            & 270            & 114            & 63            \\
\multicolumn{1}{c|}{}                                  & Memory (MiB)    & 22,126         & 17,012         & 12,262        & 19,716         & 15,422         & 11,420        \\ \hline
\multicolumn{1}{c|}{\multirow{2}{*}{128 - 10 - 64}}    & Time (min)      & 285            & 120            & 66            & 270            & 114            & 63            \\
\multicolumn{1}{c|}{}                                  & Memory (MiB)    & 22,146         & 17,016         & 12,260        & 19,736         & 15,426         & 11,420        \\ \hline
\end{tabular}
\end{table}


\clearpage

\end{CJK}

\end{document}